\pdfoutput=1

\documentclass[11pt]{article}

\usepackage{acl}

\usepackage{times}
\usepackage{latexsym}

\usepackage[T1]{fontenc}

\usepackage[utf8]{inputenc}

\usepackage{microtype}
\usepackage{starfont}
\usepackage{amsmath}
\usepackage{bm}

\usepackage[most]{tcolorbox}
\usepackage{enumitem}

\usepackage{blindtext}
\usepackage{subcaption}
\usepackage{multirow}
\usepackage{booktabs}
\usepackage{colortbl}
\usepackage{inconsolata}
\usepackage{arydshln}

\newcommand{\postspace}{\vskip -3mm}
\newcommand{\minipostspace}{\vskip -1mm}

\newcommand{\money}[0]{\includegraphics[width=.02\textwidth]{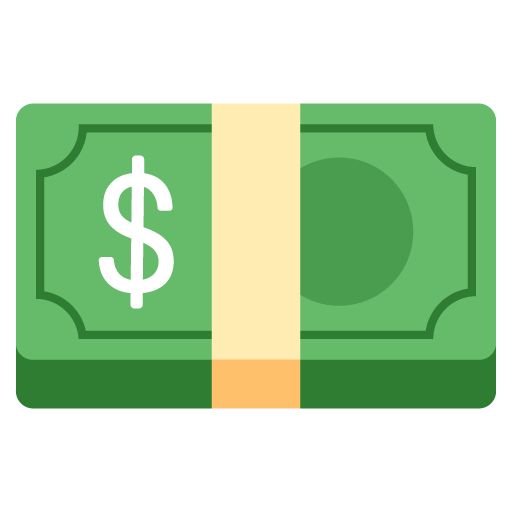}}

\definecolor{bblue}{HTML}{2E91E5}
\definecolor{ppink}{HTML}{E15F99}
\definecolor{ggreen}{HTML}{1CA71C}
\definecolor{rred}{HTML}{FB0D0D}

\title{Less is More for Long Document Summary Evaluation by LLMs}

\author{Yunshu Wu\thanks{~~Equal contribution}~~\thanks{~~The work was done when Yunshu Wu was a research intern at Megagon Labs.} \\
  University of California Riverside\\
  \texttt{ywu380@ucr.edu} \\\And
  Hayate Iso\footnotemark[1] \\
  Megagon Labs \\
  \texttt{hayate@megagon.ai} \\ \AND
  Pouya Pezeshkpour \\
  Megagon Labs \\
  \texttt{pouya@megagon.ai} \\ \And
  Nikita Bhutani \\
  Megagon Labs \\
  \texttt{nikita@megagon.ai} \\ \And
  Estevam Hruschka \\
  Megagon Labs \\
  \texttt{estevam@megagon.ai} \\
}

\begin{document}
\maketitle
\begin{abstract}
Large Language Models (LLMs) have shown promising performance in summary evaluation tasks, yet they face challenges such as high computational costs and the \textit{Lost-in-the-Middle} problem where important information in the middle of long documents is often overlooked. To address these issues, this paper introduces a novel approach, Extract-then-Evaluate, which involves extracting key sentences from a long source document and then evaluating the summary by prompting LLMs. The results reveal that the proposed method not only significantly reduces evaluation costs but also exhibits a higher correlation with human evaluations. Furthermore, we provide practical recommendations for optimal document length and sentence extraction methods, contributing to the development of cost-effective yet more accurate methods for LLM-based text generation evaluation.\footnote{The code is available at \url{https://github.com/megagonlabs/llm-longeval}}

\end{abstract}

\section{Introduction}
The evaluation of text generation plays a crucial role in the development of high-quality text generation systems~\cite{celikyilmaz2021evaluation}. However, the alignment of automatic evaluation metrics with human judgment remains a challenging task~\cite{bhandari-etal-2020-evaluating,fabbri2021summeval}. Recently, large language models (LLMs) have shown promising results in this regard~\cite{chiang-lee-2023-large,liu2023geval,fu2023gptscore}, demonstrating a strong correlation with human evaluations.
Despite their effectiveness, they face challenges such as high computational cost and the \textit{Lost-in-the-middle} problem~\cite{liu2023lost} where important information in the middle of long documents is often overlooked for long document summary evaluation. %
\begin{figure}[!t]
  \centering
  \includegraphics[width=\linewidth]{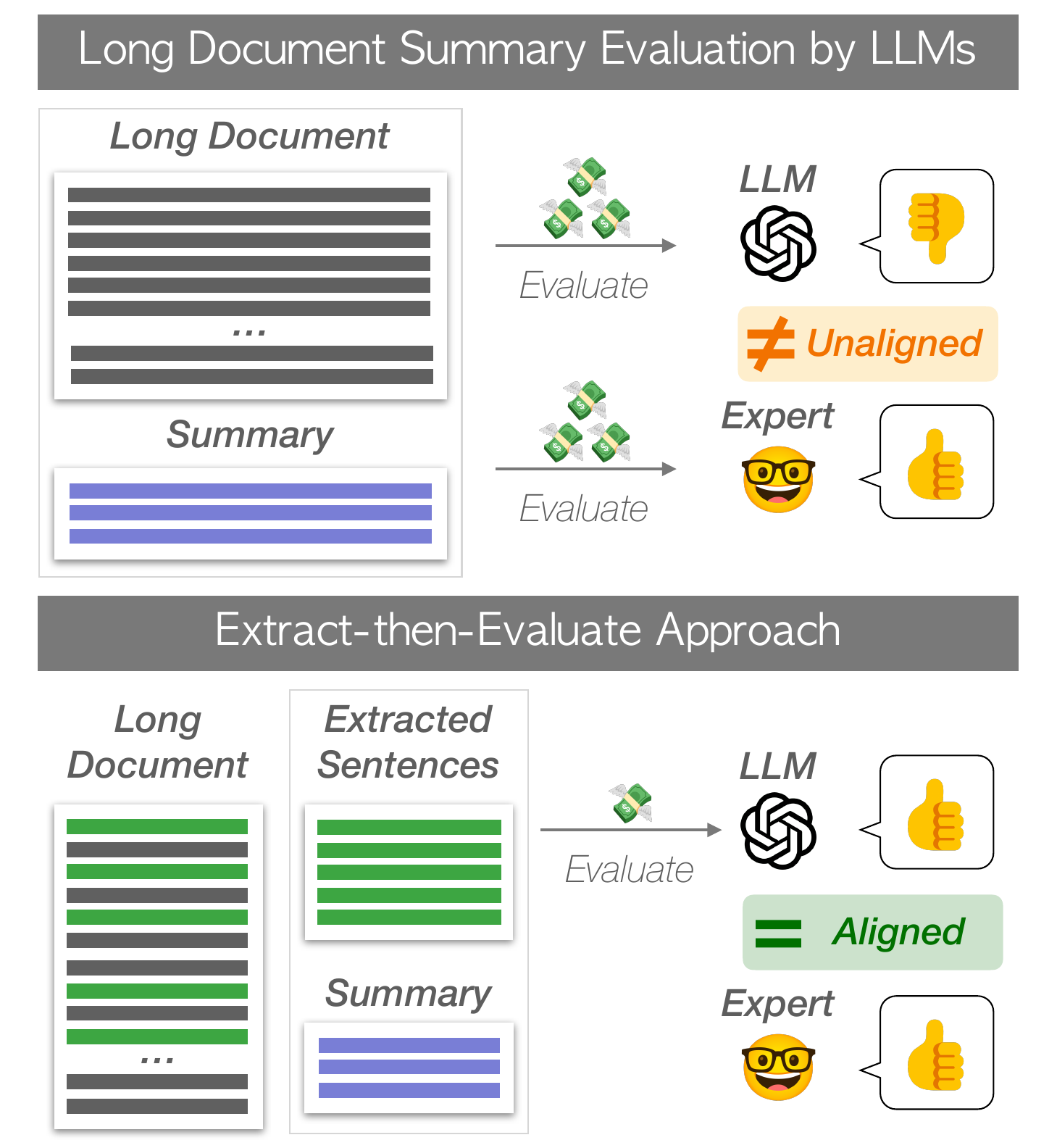}
  \caption{Overview of the long document summary evaluation task by LLMs. Evaluating long document summaries by LLMs is expensive and shows limited alignment with human evaluations. This study demonstrates that extracting important sentences for evaluation in advance not only reduces evaluation costs but also exhibits better alignment with human evaluations.}
  \label{fig:fig1}
  \postspace
  \minipostspace
\end{figure}

In this paper, we propose a simple yet effective approach to address these issues, which we refer to as the Extract-then-Evaluate. This method involves extracting important sentences from a long source document and concatenating them until the extracted document reaches a pre-defined length. Then, we evaluate the quality of the summary with regard to the extracted document using LLMs. We experiment with various sentence extraction methods---covering both matching- and model-based approaches---including LEAD, ROUGE, BERTScore, and NLI, and evaluate their performance on arXiv, GovReport, PubMed, and SQuALITY datasets~\cite{koh-etal-2022-far,krishna-etal-2023-longeval}.

Our contributions are as follows: 
\begin{itemize}[noitemsep]
    \item Develops cost-effective and efficient methods for text generation evaluation.
    \item Reduces evaluation costs and exhibits a higher correlation with human evaluations.
    \item Provides practical recommendations for optimal document length and sentence extraction methods.
\end{itemize}

\section{Methods}
\label{sec:method}

Summarization evaluation metrics assign a rating $\hat{s}$ to a model-generated summary $\hat{y}$. The higher the correlation $corr(\hat{s}, s)$ between this score $\hat{s}$ and the human judgment score $s$,  the better the evaluation metric is. 
To assign a rating $\hat{s}$, existing studies use either the reference summary $y$ or the input document $x$, as well as the generated summary $\hat{y}$.

To use LLMs as evaluators, previous approaches commonly use the model-generated summaries $\hat{y}$, and the source document $x$ as inputs, where $\hat{s} = f(x, \hat{y})$, but the Extract-then-Evaluate method comprises two steps to use LLMs as illustrated in Figure~\ref{fig:fig1}:
(1) Extract important sentences for summary evaluation from the long source document $x$ until it reaches the pre-defined length $N$, and compose a short but information-dense document $x'$.
(2) Evaluate the quality of the summary $\hat{y}$ by prompting LLMs~\cite{liu2023geval}. 
Design prompts \footnote{All prompts used are listed in the Appendix.} that can take both the extracted source document $x'$ and summary $\hat{y}$ as inputs and generate a rating scale $s$ as output: $\hat{s} = f(g_{extract}(x), \hat{y})$ 

To extract sentences, we considered the following approaches:
\begin{itemize}[noitemsep]
    \item \textbf{LEAD}: Extract the first $N$ tokens from $x$. 
    This is considered a strong baseline for extractive summarization~\cite{see-etal-2017-get}.
    \item \textbf{ROUGE}: Extract sentences from $x$ that maximize recall of ROUGE score~\cite{lin-2004-rouge} with $\hat{y}$ until it reaches $N$ tokens.\footnote{\url{https://github.com/Diego999/py-rouge}}
    \item \textbf{BERTScore}: Extract sentences as in ROUGE, but use the recall of BERTScore~\cite{zhang2019bertscore} as the criteria.
    \item \textbf{NLI}: Extract sentences that are entailed or contradicted by each sentence in $\hat{y}$ as premises using NLI models~\cite{reimers-gurevych-2019-sentence} until it reaches $N$ tokens. This process aims to extract sentences that are semantically relevant to the summary.
\end{itemize}
The source document is divided into sentences; then, important sentences are extracted based on the criteria above; if the extracted sentences reach the predefined length limit, they are rearranged to match the order in the source document.

\begin{table}[!t]
    \centering
    \small
    \resizebox{\columnwidth}{!}{
    \begin{tabular}{ccccc}
        \toprule
         &  &\textbf{\#instance}   &\textbf{Document avg length}    &\textbf{Summary avg length}\\
         \midrule
         &\textbf{arXiv}        & 204 & 5723 & 178 \\
         &\textbf{GovReport}    & 204 & 8553 & 500 \\
         &\textbf{PubMed}       & 40  & 7333 & 403 \\
         &\textbf{SQuALITY}     & 40  & 4331 & 236 \\
         \bottomrule
    \end{tabular}
    }
    \caption{Dataset statistics. The document and summary length are the average number of BPE tokens using the GPT-4 tokenizer.}
    \label{tab:stats}
\end{table}
\begin{table*}[!t]
    \centering
    \small
    \resizebox{\textwidth}{!}{
    \begin{tabular}{lcccccccccccccccccc}
        \toprule
        & \multicolumn{6}{c}{\textbf{Consistency}} & \multicolumn{6}{c}{\textbf{Relevance}} & \multicolumn{6}{c}{\textbf{Faithfulness}}\\
        & \multicolumn{3}{c}{\textbf{arXiv}} & \multicolumn{3}{c}{\textbf{GovReport}} & \multicolumn{3}{c}{\textbf{arXiv}} & \multicolumn{3}{c}{\textbf{GovReport}} & \multicolumn{3}{c}{\textbf{PubMed}} & \multicolumn{3}{c}{\textbf{SQuALITY}} \\
        \cmidrule(lr){2-4} \cmidrule(lr){5-7} \cmidrule(lr){8-10} \cmidrule(lr){11-13} \cmidrule(lr){14-16} \cmidrule(lr){17-19} 
        \textbf{Methods} & $r$ & $\rho$ & \money & $r$ & $\rho$ & \money & $r$ & $\rho$ & \money & $r$ & $\rho$ & \money & $r$ & $\rho$ & \money & $r$ & $\rho$ & \money \\\midrule
        \multicolumn{19}{c}{\textit{Reference-based metrics}}\\
        \midrule
        ROUGE-1 & -0.08 & -0.13 & - & -0.12 & -0.11 & - & 0.29 & 0.25 & - & 0.53 & 0.52 & - & 0.32 & 0.30 & - & -0.33 & -0.13 & -\\
        BERTScore & -0.09 & -0.10 & - & 0.00 & -0.04 & - & 0.22 & 0.18 & - & 0.38 & 0.38 & - & 0.49 & 0.49 & - & -0.12 & 0.02 & -\\
        BARTScore & 0.32 & 0.36 & - & 0.51 & 0.48 & - & 0.00 & 0.03 & - & 0.18 & 0.24 & - & 0.49 & 0.47 & - & -0.06 & -0.17 & -\\
        \midrule
        \multicolumn{19}{c}{\textit{Reference-free metrics}}\\
        \midrule
        FactCC & 0.22 & 0.19 & - & 0.28 & 0.27 & - & 0.13 & 0.13 & - & 0.05 & 0.04 & -  & -0.09 & -0.14 & - & 0.13 & 0.14 & -  \\
        SummaC & 0.32 & 0.32 & - & 0.39 & 0.38 & - & 0.09 & 0.08 & - & 0.05 & 0.04 & - & 0.51 & 0.55 & - & 0.18 & 0.24 & -\\
        \midrule
        \multicolumn{19}{c}{\textit{Reference-free metrics with LLM} (ours)}\\
        \midrule
        Full document & 0.61 & 0.46 & \$0.15 & 0.33 & 0.34 & \$0.10 & 0.58 & 0.52 & \$0.15 & 0.12 & 0.11 & \$0.10 & 0.64 & 0.70 & \$0.11 & 0.51 & 0.38 & \$0.14 \\
        Best extraction & 0.71 & 0.50 & \$0.05 & 0.62 & 0.60 & \$0.09 & 0.63 & 0.58 & \$0.07 & 0.36 & 0.40 & \$0.07 & 0.76 & 0.80 & \$0.07 & 0.85 & 0.81 & \$0.04\\
        Pareto efficient & 0.71 & 0.50 & \$0.05 & 0.60 & 0.61 & \$0.05 & 0.55 & 0.48 & \$0.04 & 0.37 & 0.37 & \$0.05 & 0.75 & 0.75 & \$0.05 & 0.85 & 0.81 & \$0.04\\
        \bottomrule
    \end{tabular}
    }
    \caption{Results for Pearson correlation ($r$), Spearman correlation ($\rho$), and the average evaluation cost per instance ($\money$) indicate that extracting important sentences before evaluation (Best extraction) can yield a higher correlation. %
    Even under a limited budget (Pareto efficient), these results show comparable or even higher correlations compared to the full document setting, with lower costs. We have highlighted each selected point in Table~\ref{tab:full_table} in the Appendix.}
    \label{tab:main_result}
    \postspace
    \minipostspace
\end{table*}
\section{Experiments}
\subsection{Settings}
This study meta-evaluates automatic evaluation metrics for summarization by assessing their alignment with human judgment. Specifically, each metric assigns a numerical score to the model-generated summary and measures its Pearson correlation $r$ and Spearman's rank correlation $\rho$ with the human evaluation score to measure the alignment. We also calculated the average evaluation cost of using LLMs to investigate the efficiency of our method to see how much we can save with our method.\footnote{Calculated as \$0.03 per 1k tokens of input.}
For the meta-evaluation, we used the following datasets:
 \textbf{arXiv}~\cite{cohan2018discourse} and \textbf{GovReport}~\cite{huang2021efficient}, scientific and general domain of summarization datasets, respectively, with human evaluations of \textbf{Consistency} and \textbf{Relevance} collected by \citet{koh-etal-2022-far}. 
 \textbf{PubMed}~\cite{cohan2018discourse} and \textbf{SQuALITY}~\cite{wang2022squality}, biomedical science and story domain of summarization datasets, with human evaluations of \textbf{Faithfullness} collected by \citet{krishna-etal-2023-longeval}.\footnote{We found an issue in the original evaluation, so the baseline correlation such as ROUGE-1 is inconsistent with the original paper. Please refer to the Appendix for more details.} We used fine-grained faithfulness scores as human judgments.
Table~\ref{tab:stats} shows the statistics of the datasets. 

\label{sec:result}

\subsection{Implementation Details}
We used GPT-4~\cite{openai2023gpt-4} as our evaluator~\cite{liu2023geval}.\footnote{\texttt{gpt-4-0613} checkpoint is used. See Appendix C for reasons to use GPT4.} As described in \S\ref{sec:method}, we design prompts based on the definition of each evaluation criterion and derive rating scales that evaluate the summary with deterministic predictions.\footnote{This setting is slightly different from that of \citet{liu2023geval}; more details in the Appendix.} Note that at the time of submission, access to GPT4 with 32k was not permitted, so if the prompt was longer 8k tokens, we truncated the source document $x$ to meet the length limit.

For sentence extraction, we experimented with 128, 256, 512, 768, 1024, 1536, 2048, and 4096 tokens, as the length limit $N$ of the extracted source document. For the ROUGE-based sentence extraction, we used recall of ROUGE-1, ROUGE-2, and the sum of them (ROUGE-1+2). For the BERTScore, we used DeBERTa-Large model~\cite{he2021deberta} fine-tuned on MNLI~\cite{williams-etal-2018-broad}.\footnote{\url{https://huggingface.co/microsoft/deberta-large-mnli}} For the NLI, we used DeBERTa-base model fine-tuned on SNLI~\cite{bowman-etal-2015-large} and MNLI~\cite{williams-etal-2018-broad}.\footnote{\url{https://huggingface.co/cross-encoder/nli-deberta-v3-base}}

\subsection{Baselines}

For the baseline, we use two groups of metrics: reference-based and reference-free. For the reference-based metrics, we use ROUGE-1 F1~\cite{lin-2004-rouge}, BERTScore~\cite{zhang2019bertscore}, and BARTScore~\cite{yuan2021bartscore}. For the reference-free metrics, we use FactCC~\cite{kryscinski2019evaluating}, and SummaC~\cite{laban2022summac}. Also, we use the LLM-based evaluation without sentence extraction as a baseline (\textit{Full document}).

\begin{figure*}[!t]
    \centering
    \includegraphics[width=\linewidth]{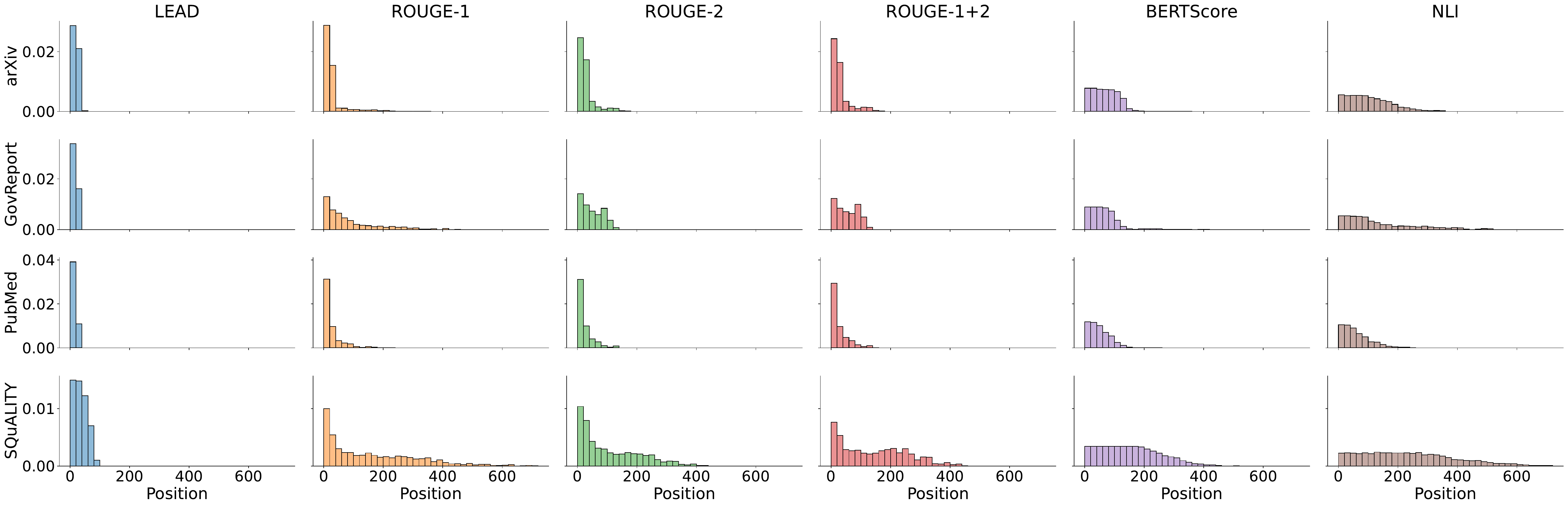}
    \caption{Distribution of sentence positions extracted by different methods. For the scientific domain, ROUGE-based methods tend to extract sentences positioned primarily at the beginning of documents. Conversely, for the general domain, ROUGE-based methods tend to choose sentences from throughout the document. Also, model-based approaches, BERTScore and NLI, tend to extract sentences from diverse locations, regardless of the dataset.}
    \label{fig:position}
    \postspace
\end{figure*}
\begin{figure*}[!t]
    \centering
    \includegraphics[width=\linewidth]{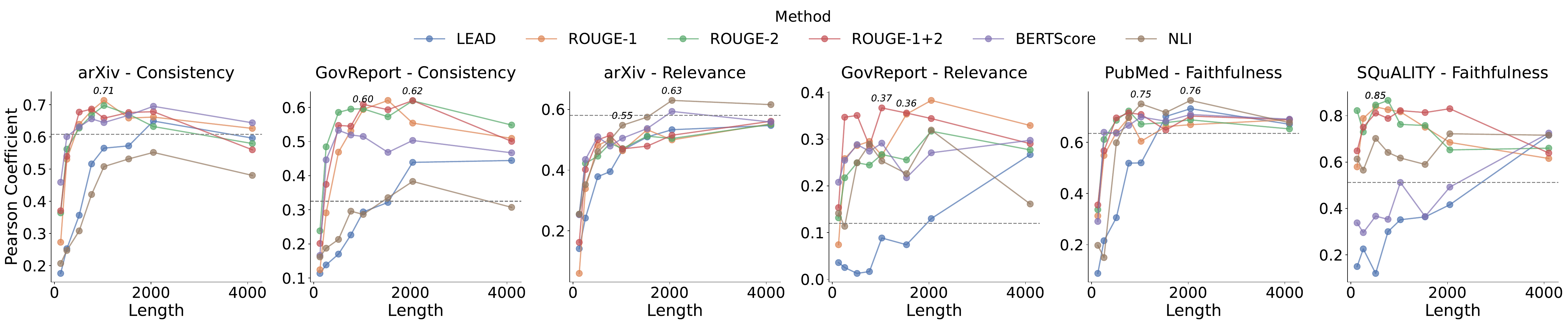}
    \caption{Relationship between document length and Pearson correlation shows the highest correlation at 1000-2000 tokens. For the scientific domain, important information is typically concentrated at the beginning (i.e., introduction). In such cases, LEAD performs comparably well. However, in the general domain, important information is more distributed throughout the document, and thus LEAD performs significantly worse than the others.}
    \label{fig:pearson}
    \postspace
\end{figure*}

\subsection{Results} 

Due to space constraints, we only provide results for two of our variations in Table~\ref{tab:main_result}: \textit{Best extraction}, yielding the highest correlation among all variations, and \textit{Pareto efficient}, which is a cost-effective approach, offering the highest correlation with the input extracted source document length under 1024 tokens. Results for all variations are shown in Table~\ref{tab:full_table} in the Appendix.

First, LLM mostly showed a significant improvement in correlation with human judgment compared to the non-LLM baselines. However, the evaluation costs definitely increased due to the entire prompt length (Full document).

Next, we observed that extracting information from the source document and then evaluating it not only lowers costs but also improves performance (Best Extraction). This could be attributed to the \textit{Lost-in-the-middle}~\cite{liu2023lost}, where LLMs struggle to efficiently use important information in the middle of long documents. In other words, LLMs would better understand shorter but more informative documents for evaluation.
Note that this observation is not limited to the best extraction setting; we have observed a trend where performance increases as the size of the document decreases.

Finally, even when evaluated on a limited budget, we confirmed comparable performance to the highest performance settings (Pareto Efficient). Specifically, for the consistency of GovReport data, our approach demonstrated similar performance to the best extraction option while reducing costs by half.

\section{Discussion}

\paragraph{How are extracted sentences distributed?}
We analyzed the positions of sentences extracted by each method. Figure~\ref{fig:position} displays the distribution of sentence positions when limiting the length to 1024 tokens. 
For the scientific domain (i.e., arXiv and PubMed), ROUGE-based methods tend to extract sentences from positions similar to the LEAD, suggesting that important information is mostly located at the beginning of these documents.

In contrast, for the general domain (i.e., GovReport and SQuALITY), ROUGE-based methods tend to extract sentences not only from the beginning but also from various positions throughout documents, indicating that important information might be distributed throughout documents. 
Meanwhile, model-based methods (i.e., BERTScore and NLI) extract sentences from various positions within the document, regardless of the dataset.

\paragraph{How long is the optimal document length?}
Figure~\ref{fig:pearson} shows the relationship between Pearson correlation and the length of documents for various datasets and evaluation criteria. The dashed lines correspond to the Full document setting. 
We observed a strong correlation within the document length range of 1000 to 2000 tokens across all datasets. Notably, extracted documents should generally be longer than the summaries, while long documents pose the \textit{Lost-in-the-Middle} challenges for LLMs~\cite{liu2023lost}, causing the correlation curves to initially rise and then decline.

\paragraph{Which sentence extraction method is the best?}
As shown in Figure~\ref{fig:pearson} (more detailed numbers can be found in Table~\ref{tab:full_table} in the Appendix), the best extraction settings differ for each data and evaluation criteria: LEAD consistently shows a lower correlation than the other methods, while the BERTScore and NLI are mixed across data and criteria. However, the ROUGE-based methods consistently show high correlations regardless of data and criteria.

\paragraph{Practical Recommendations:}
To summarize the discussion above, we offer the following recommendations:
(1) Prompting the LLM demonstrates a strong correlation with human judgment in summary evaluation, although it's not imperative to utilize the entire source document if it's too long.
(2) Our experiments indicate that the source document's length should ideally range from 1000 to 2000 tokens, and it should surpass the length of the summary.
(3) To extract sentences for evaluation, the ROUGE-based method proves to be a straightforward yet highly effective approach.

\section{Conclusion}
In this study, we proposed the Extract-then-Evaluate method for evaluating long document summaries using LLMs. Our findings demonstrated that this approach not only reduces evaluation costs but also aligns more closely with human evaluations compared to existing automatic metrics. Furthermore, we provided practical recommendations for optimal document length and sentence extraction methods, contributing to the development of more efficient and cost-effective methods for text generation evaluation using LLMs.

\section*{Limitations}
While our method achieves superior performance, it still suffers from several limitations. Previous works \citep{liu2023geval,deutsch-etal-2022-limitations} suggest that LLM-based evaluators introduce bias toward model-generated text, affecting their reliability to assess the quality of summaries fairly. 

In this work, we mainly focus on one LLM-based evaluator utilizing GPT-4 \& GPT-3.5 due to our limited budget and computational resources. Also, we rely on correlation with human annotations to evaluate the quality of metrics, which is shown to be not very reliable specifically for long document summarization \citep{krishna-etal-2023-longeval}. Further investigation of the Extract-then-Evaluate impact on other LLM-based evaluators and introduction of better evaluation methodology remains an open venue for future works

\bibliography{custom}

\begin{thebibliography}{29}
\expandafter\ifx\csname natexlab\endcsname\relax\def\natexlab#1{#1}\fi

\bibitem[{Adams et~al.(2023)Adams, Zucker, and Elhadad}]{adams2023meta}
Griffin Adams, Jason Zucker, and No{\'e}mie Elhadad. 2023.
\newblock A meta-evaluation of faithfulness metrics for long-form hospital-course summarization.
\newblock \emph{arXiv preprint arXiv:2303.03948}.

\bibitem[{Beltagy et~al.(2020)Beltagy, Peters, and Cohan}]{beltagy2020longformer}
Iz~Beltagy, Matthew~E Peters, and Arman Cohan. 2020.
\newblock \href {https://arxiv.org/abs/2004.05150} {Longformer: The long-document transformer}.
\newblock \emph{arXiv preprint arXiv:2004.05150}.

\bibitem[{Bhandari et~al.(2020)Bhandari, Gour, Ashfaq, Liu, and Neubig}]{bhandari-etal-2020-evaluating}
Manik Bhandari, Pranav~Narayan Gour, Atabak Ashfaq, Pengfei Liu, and Graham Neubig. 2020.
\newblock \href {https://doi.org/10.18653/v1/2020.emnlp-main.751} {Re-evaluating evaluation in text summarization}.
\newblock In \emph{Proceedings of the 2020 Conference on Empirical Methods in Natural Language Processing (EMNLP)}, pages 9347--9359, Online. Association for Computational Linguistics.

\bibitem[{Bowman et~al.(2015)Bowman, Angeli, Potts, and Manning}]{bowman-etal-2015-large}
Samuel~R. Bowman, Gabor Angeli, Christopher Potts, and Christopher~D. Manning. 2015.
\newblock \href {https://doi.org/10.18653/v1/D15-1075} {A large annotated corpus for learning natural language inference}.
\newblock In \emph{Proceedings of the 2015 Conference on Empirical Methods in Natural Language Processing}, pages 632--642, Lisbon, Portugal. Association for Computational Linguistics.

\bibitem[{Celikyilmaz et~al.(2020)Celikyilmaz, Clark, and Gao}]{celikyilmaz2021evaluation}
Asli Celikyilmaz, Elizabeth Clark, and Jianfeng Gao. 2020.
\newblock \href {https://arxiv.org/abs/2006.14799} {Evaluation of text generation: A survey}.
\newblock \emph{arXiv preprint arXiv:2006.14799}.

\bibitem[{Chiang and Lee(2023)}]{chiang-lee-2023-large}
Cheng-Han Chiang and Hung-yi Lee. 2023.
\newblock \href {https://doi.org/10.18653/v1/2023.acl-long.870} {Can large language models be an alternative to human evaluations?}
\newblock In \emph{Proceedings of the 61st Annual Meeting of the Association for Computational Linguistics (Volume 1: Long Papers)}, pages 15607--15631, Toronto, Canada. Association for Computational Linguistics.

\bibitem[{Cohan et~al.(2018)Cohan, Dernoncourt, Kim, Bui, Kim, Chang, and Goharian}]{cohan2018discourse}
Arman Cohan, Franck Dernoncourt, Doo~Soon Kim, Trung Bui, Seokhwan Kim, Walter Chang, and Nazli Goharian. 2018.
\newblock \href {https://doi.org/10.18653/v1/N18-2097} {A discourse-aware attention model for abstractive summarization of long documents}.
\newblock In \emph{Proceedings of the 2018 Conference of the North {A}merican Chapter of the Association for Computational Linguistics: Human Language Technologies, Volume 2 (Short Papers)}, pages 615--621, New Orleans, Louisiana. Association for Computational Linguistics.

\bibitem[{Dao et~al.(2022)Dao, Fu, Ermon, Rudra, and R\'{e}}]{dao2022flash}
Tri Dao, Dan Fu, Stefano Ermon, Atri Rudra, and Christopher R\'{e}. 2022.
\newblock \href {https://proceedings.neurips.cc/paper_files/paper/2022/file/67d57c32e20fd0a7a302cb81d36e40d5-Paper-Conference.pdf} {Flashattention: Fast and memory-efficient exact attention with io-awareness}.
\newblock In \emph{Advances in Neural Information Processing Systems}, volume~35, pages 16344--16359. Curran Associates, Inc.

\bibitem[{Deutsch et~al.(2022)Deutsch, Dror, and Roth}]{deutsch-etal-2022-limitations}
Daniel Deutsch, Rotem Dror, and Dan Roth. 2022.
\newblock \href {https://doi.org/10.18653/v1/2022.emnlp-main.753} {On the limitations of reference-free evaluations of generated text}.
\newblock In \emph{Proceedings of the 2022 Conference on Empirical Methods in Natural Language Processing}, pages 10960--10977, Abu Dhabi, United Arab Emirates. Association for Computational Linguistics.

\bibitem[{Fabbri et~al.(2021)Fabbri, Kry{\'s}ci{\'n}ski, McCann, Xiong, Socher, and Radev}]{fabbri2021summeval}
Alexander~R. Fabbri, Wojciech Kry{\'s}ci{\'n}ski, Bryan McCann, Caiming Xiong, Richard Socher, and Dragomir Radev. 2021.
\newblock \href {https://doi.org/10.1162/tacl_a_00373} {{S}umm{E}val: Re-evaluating summarization evaluation}.
\newblock \emph{Transactions of the Association for Computational Linguistics}, 9:391--409.

\bibitem[{Fu et~al.(2023)Fu, Ng, Jiang, and Liu}]{fu2023gptscore}
Jinlan Fu, See-Kiong Ng, Zhengbao Jiang, and Pengfei Liu. 2023.
\newblock \href {https://arxiv.org/abs/2302.04166} {{GPTS}core: Evaluate as you desire}.
\newblock \emph{arXiv preprint arXiv:2302.04166}.

\bibitem[{Gu et~al.(2022)Gu, Goel, and Re}]{gu2022efficiently}
Albert Gu, Karan Goel, and Christopher Re. 2022.
\newblock \href {https://openreview.net/forum?id=uYLFoz1vlAC} {Efficiently modeling long sequences with structured state spaces}.
\newblock In \emph{International Conference on Learning Representations}.

\bibitem[{He et~al.(2021)He, Liu, Gao, and Chen}]{he2021deberta}
Pengcheng He, Xiaodong Liu, Jianfeng Gao, and Weizhu Chen. 2021.
\newblock \href {https://openreview.net/forum?id=XPZIaotutsD} {{DeBERTa: Decoding-enhanced BERT with Disentangled Attention}}.
\newblock In \emph{International Conference on Learning Representations}.

\bibitem[{Huang et~al.(2021)Huang, Cao, Parulian, Ji, and Wang}]{huang2021efficient}
Luyang Huang, Shuyang Cao, Nikolaus Parulian, Heng Ji, and Lu~Wang. 2021.
\newblock \href {https://doi.org/10.18653/v1/2021.naacl-main.112} {Efficient attentions for long document summarization}.
\newblock In \emph{Proceedings of the 2021 Conference of the North American Chapter of the Association for Computational Linguistics: Human Language Technologies}, pages 1419--1436, Online. Association for Computational Linguistics.

\bibitem[{Jiang et~al.(2023)Jiang, Sablayrolles, Mensch, Bamford, Chaplot, de~las Casas, Bressand, Lengyel, Lample, Saulnier, Lavaud, Lachaux, Stock, Scao, Lavril, Wang, Lacroix, and Sayed}]{jiang2023mistral}
Albert~Q. Jiang, Alexandre Sablayrolles, Arthur Mensch, Chris Bamford, Devendra~Singh Chaplot, Diego de~las Casas, Florian Bressand, Gianna Lengyel, Guillaume Lample, Lucile Saulnier, Lélio~Renard Lavaud, Marie-Anne Lachaux, Pierre Stock, Teven~Le Scao, Thibaut Lavril, Thomas Wang, Timothée Lacroix, and William~El Sayed. 2023.
\newblock \href {http://arxiv.org/abs/2310.06825} {Mistral 7b}.

\bibitem[{Koh et~al.(2022)Koh, Ju, Zhang, Liu, and Pan}]{koh-etal-2022-far}
Huan~Yee Koh, Jiaxin Ju, He~Zhang, Ming Liu, and Shirui Pan. 2022.
\newblock \href {https://doi.org/10.18653/v1/2022.emnlp-main.172} {How far are we from robust long abstractive summarization?}
\newblock In \emph{Proceedings of the 2022 Conference on Empirical Methods in Natural Language Processing}, pages 2682--2698, Abu Dhabi, United Arab Emirates. Association for Computational Linguistics.

\bibitem[{Krishna et~al.(2023)Krishna, Bransom, Kuehl, Iyyer, Dasigi, Cohan, and Lo}]{krishna-etal-2023-longeval}
Kalpesh Krishna, Erin Bransom, Bailey Kuehl, Mohit Iyyer, Pradeep Dasigi, Arman Cohan, and Kyle Lo. 2023.
\newblock \href {https://aclanthology.org/2023.eacl-main.121} {{L}ong{E}val: Guidelines for human evaluation of faithfulness in long-form summarization}.
\newblock In \emph{Proceedings of the 17th Conference of the European Chapter of the Association for Computational Linguistics}, pages 1650--1669, Dubrovnik, Croatia. Association for Computational Linguistics.

\bibitem[{Kryscinski et~al.(2020)Kryscinski, McCann, Xiong, and Socher}]{kryscinski2019evaluating}
Wojciech Kryscinski, Bryan McCann, Caiming Xiong, and Richard Socher. 2020.
\newblock \href {https://doi.org/10.18653/v1/2020.emnlp-main.750} {Evaluating the factual consistency of abstractive text summarization}.
\newblock In \emph{Proceedings of the 2020 Conference on Empirical Methods in Natural Language Processing (EMNLP)}, pages 9332--9346, Online. Association for Computational Linguistics.

\bibitem[{Laban et~al.(2022)Laban, Schnabel, Bennett, and Hearst}]{laban2022summac}
Philippe Laban, Tobias Schnabel, Paul~N. Bennett, and Marti~A. Hearst. 2022.
\newblock \href {https://doi.org/10.1162/tacl_a_00453} {{S}umma{C}: Re-visiting {NLI}-based models for inconsistency detection in summarization}.
\newblock \emph{Transactions of the Association for Computational Linguistics}, 10:163--177.

\bibitem[{Lin(2004)}]{lin-2004-rouge}
Chin-Yew Lin. 2004.
\newblock \href {https://aclanthology.org/W04-1013} {{ROUGE}: A package for automatic evaluation of summaries}.
\newblock In \emph{Text Summarization Branches Out}, pages 74--81, Barcelona, Spain. Association for Computational Linguistics.

\bibitem[{Liu et~al.(2023{\natexlab{a}})Liu, Lin, Hewitt, Paranjape, Bevilacqua, Petroni, and Liang}]{liu2023lost}
Nelson~F Liu, Kevin Lin, John Hewitt, Ashwin Paranjape, Michele Bevilacqua, Fabio Petroni, and Percy Liang. 2023{\natexlab{a}}.
\newblock \href {https://arxiv.org/abs/2307.03172} {Lost in the middle: How language models use long contexts}.
\newblock \emph{arXiv preprint arXiv:2307.03172}.

\bibitem[{Liu et~al.(2023{\natexlab{b}})Liu, Iter, Xu, Wang, Xu, and Zhu}]{liu2023geval}
Yang Liu, Dan Iter, Yichong Xu, Shuohang Wang, Ruochen Xu, and Chenguang Zhu. 2023{\natexlab{b}}.
\newblock \href {https://doi.org/10.18653/v1/2023.emnlp-main.153} {{G}-eval: {NLG} evaluation using gpt-4 with better human alignment}.
\newblock In \emph{Proceedings of the 2023 Conference on Empirical Methods in Natural Language Processing}, pages 2511--2522, Singapore. Association for Computational Linguistics.

\bibitem[{OpenAI(2023)}]{openai2023gpt-4}
OpenAI. 2023.
\newblock \href {https://arxiv.org/abs/2303.08774} {{GPT-4 Technical Report}}.
\newblock \emph{arXiv preprint arXiv:2303.08774}.

\bibitem[{Reimers and Gurevych(2019)}]{reimers-gurevych-2019-sentence}
Nils Reimers and Iryna Gurevych. 2019.
\newblock \href {https://doi.org/10.18653/v1/D19-1410} {Sentence-{BERT}: Sentence embeddings using {S}iamese {BERT}-networks}.
\newblock In \emph{Proceedings of the 2019 Conference on Empirical Methods in Natural Language Processing and the 9th International Joint Conference on Natural Language Processing (EMNLP-IJCNLP)}, pages 3982--3992, Hong Kong, China. Association for Computational Linguistics.

\bibitem[{See et~al.(2017)See, Liu, and Manning}]{see-etal-2017-get}
Abigail See, Peter~J. Liu, and Christopher~D. Manning. 2017.
\newblock \href {https://doi.org/10.18653/v1/P17-1099} {Get to the point: Summarization with pointer-generator networks}.
\newblock In \emph{Proceedings of the 55th Annual Meeting of the Association for Computational Linguistics (Volume 1: Long Papers)}, pages 1073--1083, Vancouver, Canada. Association for Computational Linguistics.

\bibitem[{Wang et~al.(2022)Wang, Pang, Chen, Phang, and Bowman}]{wang2022squality}
Alex Wang, Richard~Yuanzhe Pang, Angelica Chen, Jason Phang, and Samuel~R. Bowman. 2022.
\newblock \href {https://doi.org/10.18653/v1/2022.emnlp-main.75} {{SQ}u{ALITY}: Building a long-document summarization dataset the hard way}.
\newblock In \emph{Proceedings of the 2022 Conference on Empirical Methods in Natural Language Processing}, pages 1139--1156, Abu Dhabi, United Arab Emirates. Association for Computational Linguistics.

\bibitem[{Williams et~al.(2018)Williams, Nangia, and Bowman}]{williams-etal-2018-broad}
Adina Williams, Nikita Nangia, and Samuel Bowman. 2018.
\newblock \href {https://doi.org/10.18653/v1/N18-1101} {A broad-coverage challenge corpus for sentence understanding through inference}.
\newblock In \emph{Proceedings of the 2018 Conference of the North {A}merican Chapter of the Association for Computational Linguistics: Human Language Technologies, Volume 1 (Long Papers)}, pages 1112--1122, New Orleans, Louisiana. Association for Computational Linguistics.

\bibitem[{Yuan et~al.(2021)Yuan, Neubig, and Liu}]{yuan2021bartscore}
Weizhe Yuan, Graham Neubig, and Pengfei Liu. 2021.
\newblock \href {https://proceedings.neurips.cc/paper_files/paper/2021/file/e4d2b6e6fdeca3e60e0f1a62fee3d9dd-Paper.pdf} {Bartscore: Evaluating generated text as text generation}.
\newblock In \emph{Advances in Neural Information Processing Systems}, volume~34, pages 27263--27277. Curran Associates, Inc.

\bibitem[{Zhang et~al.(2020)Zhang, Kishore, Wu, Weinberger, and Artzi}]{zhang2019bertscore}
Tianyi Zhang, Varsha Kishore, Felix Wu, Kilian~Q. Weinberger, and Yoav Artzi. 2020.
\newblock \href {https://openreview.net/forum?id=SkeHuCVFDr} {Bertscore: Evaluating text generation with bert}.
\newblock In \emph{International Conference on Learning Representations}.

\end{thebibliography}
\bibliographystyle{acl_natbib}

\onecolumn
\appendix
\section{List of the Prompts}
\label{sec:prompts}

\begin{figure}[!ht]
    \begin{tcolorbox}[fontupper=\ttfamily, title={\small \texttt{Consistency}}]
    \scriptsize
    \# Instruction:\\
    Below is an instruction for evaluating the consistency of the generated summary to the source article. Consistency measures whether a candidate summary is factually consistent with the source. The goal is to score consistency on a scale of 1-5, with 1 being completely inconsistent and 5 being completely consistent.\\
    
    Please consider the following seven types of errors while performing the evaluation: i) predicate in summary inconsistent with source, ii) primary arguments or its attributes are wrong, iii) predicate’s circumstantial information is wrong, iv) co-reference error, v) multiple sentences linked incorrectly, vi) out of article error and vii) unreadable sentence(s) due to grammatical errors.\\
    
    \# Evaluation Criteria:
    \begin{enumerate}[nolistsep]
        \item Completely Inconsistent - The summary contains multiple factual errors or inaccuracies in relation to the source article.
        \item Mostly Inconsistent - The summary contains several factual errors but retains some accurate information from the source.
        \item Somewhat Consistent - The summary contains a mix of accurate and inaccurate information. Factual errors are present but not overwhelming.
        \item Mostly Consistent - The summary is largely accurate, with few factual errors or inaccuracies.
        \item Completely Consistent - The summary accurately represents all the information presented in the source article without any factual error.
    \end{enumerate}
    ~\\
    \# Evaluation Steps:
    \begin{enumerate}[nolistsep]
        \item Thoroughly read the source article.
        \item Carefully read the generated summary and compare it with the source article.
        \item Rate the consistency of the generated summary based on the provided types of errors using the 1-5 scale mentioned in Evaluation Criteria.
    \end{enumerate}
    ~\\
    \# Source Article:\\
    \{\{article\}\}
    \\\\
    \# Generated Summary:\\
    \{\{summary\}\}
    \\\\
    \# Evaluation Form (scores ONLY):
    \end{tcolorbox}
    \caption{The prompt used for evaluating the consistency of the summary.}
    \label{fig:consistency_prompt}
\end{figure}

\begin{figure}[!ht]
    \centering
    \begin{tcolorbox}[fontupper=\ttfamily, title={\small \texttt{Relevance}}]
    \scriptsize
    \# Instruction:\\
    Below is an instruction for evaluating the relevance of the generated summary to the source article. Relevance measures whether a summary contains the main ideas of the source. The goal is to score relevance on a scale of 1-5, with 1 being not relevant at all, and 5 being highly relevant.\\
    
    \# Evaluation Criteria:
    \begin{enumerate}[nolistsep]
        \item Not Relevant: The summary doesn’t capture any of the main ideas of the source.
        \item Barely Relevant: The summary captures very few of the main ideas of the source.
        \item Somewhat Relevant: The summary captures some, but not all, of the main ideas of the source.
        \item Mostly Relevant: The summary captures most of the main ideas of the source.
        \item Highly Relevant: The summary captures all the main ideas of the source perfectly.
    \end{enumerate}
    ~\\
    \# Evaluation Steps:
    \begin{enumerate}[nolistsep]
        \item Thoroughly read the source article.
        \item Carefully read the generated summary and compare it with the source article.
        \item Compare the main ideas captured in the summary to the main ideas from the source article.
        \item Rate the relevance of the summary based on how well it captures the main ideas from the source article using the 1-5 scale mentioned in Evaluation Criteria.
    \end{enumerate}
    ~\\
    \# Source Article:\\
    \{\{article\}\}
    \\\\
    \# Generated Summary:\\
    \{\{summary\}\}
    \\\\
    \# Evaluation Form (scores ONLY):
    \end{tcolorbox}
    \caption{The prompt used for evaluating the relevance of the summary.}
    \label{fig:relevance_prompt}
\end{figure}

\begin{figure*}[!tbp]
    \begin{tcolorbox}[fontupper=\ttfamily, title={\small \texttt{Faithfulness}}]
    \scriptsize
    \# Instruction:\\
    Below is an instruction for evaluating the faithfulness of the generated summary to the source article. Faithfulness is the absence of factual errors in the summary, where a factual error is a statement that contradicts the source article or is not directly stated, heavily implied, or logically entailed by the source article. The goal is to score faithfulness on a scale of 1-7, with 1 being unfaithful (all information is wrong) and 7 being extremely faithful (no factual errors, directly correlate to the article).\\
    
    \# Evaluation Criteria:
    \begin{enumerate}[nolistsep]
        \item Unfaithful: The summary contains no factual information from the article.
        \item Mostly Unfaithful: The summary contains very few factual information from the article.
        \item Somewhat Unfaithful: The summary contains some factual information but several are wrong or misleading.
        \item Neutral: The summary is half correct and half incorrect in terms of factual information.
        \item Somewhat Faithful: The summary contains more factual information than errors but still has noticeable mistakes.
        \item Mostly Faithful: The summary contains almost all factual information from the article with minor mistakes.
        \item Extremely Faithful: The summary contains all factual information from the article with no errors.
    \end{enumerate}
    ~\\
    \# Evaluation Steps:
    \begin{enumerate}[nolistsep]
        \item Thoroughly read the source article.
        \item Carefully read the generated summary and compare it with the source article.
        \item Carefully read the summary and compare the facts presented with the facts in the source article.
        \item Rate the faithfulness of the generated summary based on how faithfully the summary reflects the information in the source article using the 1-7 scale mentioned in Evaluation Criteria.
    \end{enumerate}
    ~\\
    \# Source Article:\\
    \{\{article\}\}
    \\\\
    \# Generated Summary:\\
    \{\{summary\}\}
    \\\\
    \# Evaluation Form (scores ONLY):
    \end{tcolorbox}
    \caption{The prompt used for evaluating the faithfulness of the summary.}
    \label{fig:faithfulness_prompt}
\end{figure*}

\clearpage

\section{Correlation performance between human ratings and model-based scoring}
\begin{table*}[!h]
\centering
\resizebox{\textwidth}{!}{

\begin{tabular}{lccccccccccccc}
\toprule
        & & \multicolumn{4}{c}{\textbf{Consistency}} & \multicolumn{4}{c}{\textbf{Relevance}} & \multicolumn{4}{c}{\textbf{Faithfulness}} \\
        & & \multicolumn{2}{c}{\textbf{arXiv}} & \multicolumn{2}{c}{\textbf{GovReport}} & \multicolumn{2}{c}{\textbf{arXiv}} & \multicolumn{2}{c}{\textbf{GovReport}} & \multicolumn{2}{c}{\textbf{PubMed}} & \multicolumn{2}{c}{\textbf{SQuALITY}}\\
        \cmidrule(lr){3-4} \cmidrule(lr){5-6} \cmidrule(lr){7-8} \cmidrule(lr){9-10} \cmidrule(lr){11-12} \cmidrule(lr){13-14} 
        \textbf{Methods} & \textbf{Length} & $r$ & $\rho$ & $r$ & $\rho$ & $r$ & $\rho$ & $r$ & $\rho$ & $r$ & $\rho$ & $r$ & $\rho$ \\\midrule
    \multirow{8}{*}{LEAD} &    128 &  0.1759 &   0.1104 &      0.1135 &   0.1075 &    0.1412 &   0.1542 &    0.0358 &   0.0249 &       0.0881 &   0.0483 &       0.1496 &   0.1234 \\
     &    256 &  0.2526 &   0.1834 &      0.1384 &   0.1261 &    0.2420 &   0.2097 &    0.0253 &   0.0221 &       0.2157 &   0.1749 &       0.2256 &   0.2995 \\
     &    512 &  0.3566 &   0.2434 &      0.1701 &   0.1340 &    0.3785 &   0.3173 &    0.0127 &   0.0064 &       0.3057 &   0.3488 &       0.1200 &   0.2246 \\
     &    768 &  0.5161 &   0.4190 &      0.2262 &   0.1917 &    0.3951 &   0.3399 &    0.0167 &   0.0248 &       0.5184 &   0.5199 &       0.3001 &   0.3646 \\
     &   1024 &  0.5650 &   0.4424 &      0.2938 &   0.2876 &    0.4657 &   0.3853 &    0.0885 &   0.0937 &       0.5199 &   0.5479 &       0.3514 &   0.3718 \\
     &   1536 &  0.5722 &   0.4940 &      0.3216 &   0.3319 &    0.5094 &   0.4242 &    0.0741 &   0.0844 &       0.7009 &   0.7336 &       0.3636 &   0.3881 \\
     &   2048 &  0.6493 &   0.5352 &      0.4390 &   0.4586 &    0.5332 &   0.4443 &    0.1300 &   0.1263 &       0.7313 &   0.7478 &       0.4162 &   0.4853 \\
     &   4096 &  0.5963 &   0.4433 &      0.4445 &   0.4413 &    0.5471 &   0.4864 &    0.2670 &   0.2883 &       0.6704 &   0.6905 &       0.7156 &   0.4996 \\\midrule
    \multirow{8}{*}{ROUGE-1} &    128 &  0.2727 &   0.2036 &      0.1242 &   0.0946 &    0.0596 &  -0.0024 &    0.0741 &   0.0687 &       0.3127 &   0.2706 &       0.5793 &   0.4068 \\
     &    256 &  0.5305 &   0.3803 &      0.2909 &   0.2767 &    0.3389 &   0.1939 &    0.2584 &   0.2406 &       0.5484 &   0.5938 &       0.7881 &   0.6592 \\
     &    512 &  0.6393 &   0.4290 &      0.4690 &   0.4581 &    0.4810 &   0.3759 &    0.2864 &   0.3109 &       0.6385 &   0.6715 &       0.8381 &   0.7709 \\
     &    768 &  0.6818 &   0.4349 &      0.5315 &   0.5302 &    0.5018 &   0.4170 &    0.2952 &   0.2932 &       0.6958 &   0.7140 &       0.8259 &   0.7279 \\
     &   1024 &  \cellcolor{ppink!30}0.7134 &  \cellcolor{ppink!30} 0.4964 &      0.5940 &   0.5785 &    0.4638 &   0.3543 &    0.2652 &   0.2961 &       0.6040 &   0.6559 &       0.8167 &   0.6936 \\
     &   1536 &  0.6586 &   0.4603 &      0.6206 &   0.5963 &    0.5332 &   0.4555 &    0.3536 &   0.3374 &       0.6613 &   0.6835 &       0.7501 &   0.5840 \\
     &   2048 &  0.6616 &   0.4676 &      0.5541 &   0.5562 &    0.4996 &   0.4250 &    0.3830 &   0.3563 &       0.6688 &   0.7110 &       0.6847 &   0.5560 \\
     &   4096 &  0.6264 &   0.4463 &      0.5094 &   0.4914 &    0.5526 &   0.4759 &    0.3293 &   0.3174 &       0.6883 &   0.7080 &       0.6154 &   0.3281 \\\midrule
    \multirow{8}{*}{ROUGE-2} &    128 &  0.3640 &   0.2426 &      0.2382 &   0.2110 &    0.2548 &   0.0628 &    0.1317 &   0.1349 &       0.3370 &   0.3906 &       0.8219 &   0.7283 \\
     &    256 &  0.5620 &   0.3608 &      0.4845 &   0.4659 &    0.4221 &   0.2972 &    0.2174 &   0.1720 &       0.6111 &   0.5874 &       0.7299 &   0.6378 \\
     &    512 &  0.6274 &   0.3864 &      0.5855 &   0.5769 &    0.4460 &   0.3334 &    0.2495 &   0.2276 &       0.6859 &   0.7119 &       \cellcolor{ppink!30}0.8461 &   \cellcolor{ppink!30}0.8067 \\
     &    768 &  0.6673 &   0.3888 &      0.5952 &   0.5781 &    0.4881 &   0.3950 &    0.2446 &   0.2799 &       0.7222 &   0.7627 &       0.8658 &   0.7526 \\
     &   1024 &  0.6975 &   0.4482 &      \cellcolor{ggreen!30}0.5959 &   \cellcolor{ggreen!30}0.6117 &    0.4712 &   0.3651 &    0.2673 &   0.3098 &       0.6708 &   0.7030 &       0.7624 &   0.6763 \\
     &   1536 &  0.6707 &   0.3924 &      0.5727 &   0.5589 &    0.5120 &   0.4198 &    0.2556 &   0.2738 &       0.6770 &   0.7108 &       0.7576 &   0.6844 \\
     &   2048 &  0.6322 &   0.4135 &      0.6194 &   0.5883 &    0.5043 &   0.4197 &    0.3171 &   0.2872 &       0.6876 &   0.7043 &       0.6524 &   0.5210 \\
     &   4096 &  0.5794 &   0.3844 &      0.5484 &   0.5230 &    0.5509 &   0.4734 &    0.2771 &   0.2545 &       0.6523 &   0.6983 &       0.6600 &   0.4149 \\\midrule
    \multirow{8}{*}{ROUGE-1+2} &    128 &  0.3705 &   0.2235 &      0.2013 &   0.1525 &    0.1618 &  -0.0189 &    0.1535 &   0.1480 &       0.3553 &   0.3485 &       0.6482 &   0.6282 \\
     &    256 &  0.5397 &   0.3581 &      0.3744 &   0.3623 &    0.4019 &   0.2792 &    0.3470 &   0.3054 &       0.5670 &   0.5980 &       0.7501 &   0.6522 \\
     &    512 &  0.6770 &   0.4224 &      0.5473 &   0.5205 &    0.4998 &   0.3954 &    0.3508 &   0.3332 &       0.6953 &   0.7095 &       0.8110 &   0.6452 \\
     &    768 &  0.6865 &   0.4310 &      0.5450 &   0.5303 &    0.5147 &   0.4219 &    0.2858 &   0.2974 &       0.7148 &   0.7441 &       0.7881 &   0.7055 \\
     &   1024 &  0.6581 &   0.4435 &      0.6091 &   0.5919 &    0.4700 &   0.3656 &    \cellcolor{ggreen!30}0.3669 &   \cellcolor{ggreen!30}0.3712 &       0.7088 &   0.7479 &       0.8218 &   0.7283 \\
     &   1536 &  0.6758 &   0.4393 &      0.5933 &   0.5891 &    0.4791 &   0.3750 &    \cellcolor{bblue!30}0.3560 &   \cellcolor{bblue!30}0.4030 &       0.6476 &   0.6774 &       0.8135 &   0.7370 \\
     &   2048 &  0.6784 &   0.4569 &      \cellcolor{bblue!30}0.6202 &   \cellcolor{bblue!30}0.6031 &    0.5150 &   0.4359 &    0.3442 &   0.3066 &       0.7024 &   0.7267 &       0.8300 &   0.7117 \\
     &   4096 &  0.5600 &   0.3681 &      0.5005 &   0.4688 &    0.5611 &   0.4866 &    0.2904 &   0.2757 &       0.6883 &   0.7143 &       0.6389 &   0.5220 \\\midrule
    \multirow{8}{*}{BERTScore} &    128 &  0.4590 &   0.3179 &      0.1662 &   0.1337 &    0.2529 &   0.0459 &    0.2078 &   0.2158 &       0.2910 &   0.3228 &       0.3379 &   0.5015 \\
     &    256 &  0.6008 &   0.3543 &      0.4464 &   0.4081 &    0.4351 &   0.3001 &    0.2547 &   0.2019 &       0.6392 &   0.6539 &       0.2959 &   0.3722 \\
     &    512 &  0.6313 &   0.4060 &      0.5330 &   0.5244 &    0.5102 &   0.3971 &    0.2885 &   0.2420 &       0.6355 &   0.6731 &       0.3669 &   0.4941 \\
     &    768 &  0.6561 &   0.4079 &      0.5193 &   0.5356 &    0.4794 &   0.3710 &    0.2742 &   0.1953 &       0.6658 &   0.6971 &       0.3532 &   0.3245 \\
     &   1024 &  0.6445 &   0.4110 &      0.5149 &   0.5099 &    0.5053 &   0.4132 &    0.2915 &   0.2334 &       0.6988 &   0.7226 &       0.5121 &   0.5310 \\
     &   1536 &  0.6673 &   0.4069 &      0.4683 &   0.4513 &    0.5372 &   0.4666 &    0.2176 &   0.2035 &       0.6825 &   0.7227 &       0.3653 &   0.4106 \\
     &   2048 &  0.6951 &   0.4468 &      0.5032 &   0.5265 &    0.5935 &   0.5268 &    0.2709 &   0.2117 &       0.7084 &   0.7403 &       0.4921 &   0.5091 \\
     &   4096 &  0.6438 &   0.5180 &      0.4670 &   0.4454 &    0.5585 &   0.4796 &    0.2976 &   0.2650 &       0.6904 &   0.7342 &       0.7250 &   0.5543 \\\midrule
    \multirow{8}{*}{NLI} &    128 &  0.2068 &   0.2044 &      0.1618 &   0.1369 &    0.2549 &   0.2815 &    0.1414 &   0.1307 &       0.1977 &   0.1966 &       0.6132 &   0.3684 \\
     &    256 &  0.2473 &   0.1840 &      0.1873 &   0.1964 &    0.3520 &   0.3060 &    0.1135 &   0.0979 &       0.1499 &   0.1500 &       0.5651 &   0.3486 \\
     &    512 &  0.3080 &   0.2241 &      0.2131 &   0.2099 &    0.4610 &   0.4122 &    0.2495 &   0.2454 &       0.5983 &   0.5765 &       0.7019 &   0.5427 \\
     &    768 &  0.4211 &   0.3288 &      0.2959 &   0.3063 &    0.4990 &   0.4276 &    0.2893 &   0.3008 &       0.6973 &   0.6756 &       0.6414 &   0.4565 \\
     &   1024 &  0.5078 &   0.3010 &      0.2864 &   0.2848 &    \cellcolor{ggreen!30}0.5479 &   \cellcolor{ggreen!30}0.4822 &    0.2533 &   0.2936 &       \cellcolor{ggreen!30}0.7500 &   \cellcolor{ggreen!30}0.7478 &       0.6175 &   0.3985 \\
     &   1536 &  0.5316 &   0.2834 &      0.3355 &   0.3486 &    0.5747 &   0.5009 &    0.2262 &   0.2520 &       0.7163 &   0.7316 &       0.5898 &   0.4783 \\
     &   2048 &  0.5518 &   0.3422 &      0.3831 &   0.4005 &    \cellcolor{bblue!30}0.6298 &   \cellcolor{bblue!30}0.5798 &    0.3195 &   0.3600 &       \cellcolor{bblue!30}0.7636 &   \cellcolor{bblue!30}0.7996 &       0.7219 &   0.5753 \\
     &   4096 &  0.4804 &   0.3111 &      0.3071 &   0.3254 &    0.6159 &   0.5676 &    0.1613 &   0.2452 &       0.6766 &   0.6759 &       0.7158 &   0.4570 \\
\bottomrule
\end{tabular}
}
\caption{All results of correlation with human evaluations. Highlighted in \colorbox{bblue!30}{blue} are the highest correlations (Best extraction), while \colorbox{ggreen!30}{green} indicates settings that achieved the highest correlations within budget constraints (i.e., 1024 tokens for source document) (Pareto Efficient), and \colorbox{ppink!30}{pink} denotes those meeting both criteria.}
\label{tab:full_table}
\end{table*}
\clearpage

\section{Correlation performance by GPT-3.5}
As an ablation study, Table~\ref{tab:full_table_gpt3.5} shows the results of experiments using GPT-3.5, a smaller model than GPT-4.
Unlike G-Eval, GPT-3.5 showed an overwhelmingly lower correlation than GPT4 in all data sets and settings, meaning that a GPT-4 scale model should be used as the backbone for long-document summary evaluation.
We also tested open LLM alternatives such as Mistral-7B~\cite{jiang2023mistral}, but we observed similar trends with GPT-3.5. 
Thus, we only utilize GPT-4 in this study.
\begin{table*}[!h]
\centering
\resizebox{\textwidth}{!}{
\begin{tabular}{lccccccccccccc}
\toprule
 & & \multicolumn{4}{c}{\textbf{Consistency}} & \multicolumn{4}{c}{\textbf{Relevance}} & \multicolumn{4}{c}{\textbf{Faithfulness}} \\
 & & \multicolumn{2}{c}{\textbf{arXiv}} & \multicolumn{2}{c}{\textbf{GovReport}} & \multicolumn{2}{c}{\textbf{arXiv}} & \multicolumn{2}{c}{\textbf{GovReport}} & \multicolumn{2}{c}{\textbf{PubMed}} & \multicolumn{2}{c}{\textbf{SQuALITY}}\\
 \cmidrule(lr){3-4} \cmidrule(lr){5-6} \cmidrule(lr){7-8} \cmidrule(lr){9-10} \cmidrule(lr){11-12} \cmidrule(lr){13-14} 
 \textbf{Methods} & \textbf{Length} & $r$ & $\rho$ & $r$ & $\rho$ & $r$ & $\rho$ & $r$ & $\rho$ & $r$ & $\rho$ & $r$ & $\rho$ \\\midrule
\multirow{8}{*}{LEAD} & 128 &-0.0631 & -0.1246 & -0.0816 &-0.0875 & 0.1558 & 0.0523 & 0.0179 & -0.0150 & 0.3237 & 0.3638 & -0.1130 & 0.0167 \\
& 256 & 0.0907 & 0.0612 & -0.0943 &-0.1975 & 0.2838 & 0.0848 & 0.0765 & 0.0680 & 0.3746 & 0.4273 & -0.0551 & 0.1174 \\
& 512 & 0.1018 & 0.0836 & 0.0304 & 0.0063 & 0.3264 & 0.1809 & -0.0144 & 0.0112 & 0.4784 & 0.4774 & -0.2493 & -0.0656 \\
& 768 & 0.1120 & 0.1282 & -0.1631 &-0.1420 & 0.3208 & 0.1279 & -0.0131 & 0.0119 & 0.4779 & 0.4929 & 0.0444 & 0.1804 \\
& 1024 & 0.1345 & 0.1924 & -0.1232 &-0.1065 & 0.3589 & 0.2247 & -0.0883 & -0.0615 & 0.5467 & 0.5365 & 0.0769 & 0.3077 \\
& 1536 & 0.0243 & 0.0510 & -0.0972 &-0.1063 & 0.4035 & 0.2878 & -0.1134 & -0.1159 & 0.4573 & 0.4729 & 0.2153 & 0.2649 \\
& 2048 & 0.0648 & 0.0944 & 0.1180 & 0.0419 & 0.3629 & 0.1862 & -0.0850 & -0.0646 & 0.4834 & 0.4387 & -0.0742 & 0.1291 \\
& 4096 & 0.1432 & 0.2804 & 0.0076 &-0.0320 & 0.4003 & 0.2877 & -0.0810 & -0.1366 & 0.4887 & 0.5235 & 0.3941 & 0.5443 \\\midrule
\multirow{8}{*}{ROUGE-1} & 128 & 0.0953 & 0.0308 & 0.1144 & 0.0270 & 0.2975 &-0.0156 & 0.0132 & 0.0197 & 0.3057 & 0.3272 & 0.1416 & 0.1791 \\
& 256 & 0.1554 & 0.1664 & -0.0514 &-0.0267 & 0.3669 & 0.2558 & 0.0992 & 0.0875 & 0.5131 & 0.5748 & 0.3521 & 0.4076 \\
& 512 & 0.1778 & 0.1719 & -0.1018 &-0.0676 & 0.3381 & 0.1484 & -0.0120 & -0.0092 & 0.5950 & 0.6350 & 0.4577 & 0.4663 \\
& 768 & 0.1025 & 0.0756 & -0.0687 &-0.0827 & 0.3907 & 0.1474 & 0.0370 & 0.0512 & 0.5308 & 0.5892 & 0.3026 & 0.3691 \\
& 1024 & 0.0466 & 0.0197 & -0.0296 &-0.0305 & 0.4263 & 0.2693 & 0.0085 & 0.0355 & 0.5364 & 0.5990 & 0.3094 & 0.2800 \\
& 1536 & 0.0091 & 0.0183 & -0.1424 &-0.1922 & 0.4150 & 0.2807 & -0.0167 & 0.0245 & 0.5344 & 0.5465 & 0.2559 & 0.3434 \\
& 2048 & 0.0582 & 0.0929 & 0.0412 &-0.0523 & 0.3718 & 0.1942 & -0.0983 & -0.0861 & 0.5765 & 0.6302 & 0.3316 & 0.3250 \\
& 4096 & 0.1276 & 0.1803 & -0.0294 &-0.0926 & 0.3365 & 0.2667 & -0.1158 & -0.1489 & 0.5377 & 0.5381 & 0.3466 & 0.3996 \\\midrule
\multirow{8}{*}{ROUGE-2} & 128 & 0.0364 & 0.0423 & 0.0024 & 0.0122 & 0.3004 & 0.0800 & 0.0241 & 0.0265 & 0.5430 & 0.5401 & 0.1911 & 0.1416 \\
& 256 & 0.1788 & 0.2386 & 0.1411 & 0.0606 & 0.3431 & 0.1536 & 0.0311 & -0.0030 & 0.5061 & 0.5506 & 0.2393 & 0.2552 \\
& 512 & 0.1457 & 0.1493 & 0.0128 & 0.0028 & 0.3525 & 0.1269 & 0.0116 & 0.0283 & 0.5243 & 0.6459 & 0.4363 & 0.5286 \\
& 768 & 0.1986 & 0.1910 & -0.0876 &-0.0379 & 0.3698 & 0.1799 & 0.0384 & 0.0608 & 0.5795 & 0.5781 & 0.4342 & 0.4749 \\
& 1024 & 0.1456 & 0.1295 & -0.0335 &-0.0578 & 0.3868 & 0.2088 & 0.0561 & 0.1093 & 0.5534 & 0.5801 & 0.2674 & 0.3082 \\
& 1536 & 0.0832 & 0.0774 & -0.0373 & 0.0298 & 0.3612 & 0.1097 & -0.0325 & -0.0142 & 0.5631 & 0.5948 & 0.3126 & 0.1937 \\
& 2048 & 0.0856 & 0.0809 & -0.0570 &-0.1089 & 0.3271 & 0.1432 & -0.0601 & -0.0584 & 0.5113 & 0.5279 & 0.2365 & 0.2271 \\
& 4096 & 0.1308 & 0.2052 & 0.0108 & 0.0160 & 0.3897 & 0.2617 & -0.1390 & -0.2079 & 0.4865 & 0.4215 & 0.4343 & 0.4465 \\\midrule
\multirow{8}{*}{ROUGE-1+2} & 128 & 0.0743 & 0.0574 & 0.0817 & 0.0436 & 0.3436 & 0.1484 & 0.0868 & 0.0550 & 0.5588 & 0.5502 & 0.3269 & 0.3056 \\
& 256 & 0.1901 & 0.2732 & 0.0833 & 0.0554 & 0.3159 & 0.1260 & 0.0922 & 0.0784 & 0.4652 & 0.4570 & 0.3900 & 0.3796 \\
& 512 & 0.1638 & 0.1769 & 0.1723 & 0.0819 & 0.3426 & 0.1366 & 0.0289 & 0.0472 & 0.5413 & 0.5490 & 0.2555 & 0.3559 \\
& 768 & 0.1467 & 0.1171 & -0.0991 &-0.0729 & 0.4152 & 0.2936 & -0.0403 & -0.0218 & 0.5379 & 0.5685 & 0.2959 & 0.3098 \\
& 1024 & 0.1211 & 0.1103 & 0.0083 &-0.0058 & 0.3679 & 0.1893 & 0.0008 & 0.0246 & 0.5615 & 0.5845 & 0.3195 & 0.3410 \\
& 1536 & 0.0772 & 0.0493 & 0.0436 & 0.0227 & 0.3998 & 0.2343 & -0.0225 & 0.0036 & 0.5691 & 0.6258 & 0.2155 & 0.2465 \\
& 2048 & 0.0499 & 0.0513 & 0.1118 & 0.0377 & 0.3657 & 0.1798 & -0.0429 & -0.0030 & 0.4922 & 0.5270 & 0.1963 & 0.3031 \\
& 4096 & 0.0663 & 0.1394 & -0.0139 &-0.0087 & 0.4393 & 0.3549 & -0.0462 & -0.0996 & 0.5561 & 0.5543 & 0.3961 & 0.4997 \\\midrule
\multirow{8}{*}{BERTScore} & 128 & 0.0528 & 0.0205 & -0.1043 &-0.1016 & 0.3069 & 0.1131 & 0.0587 & 0.0540 & 0.4424 & 0.4715 & 0.0307 & 0.1545 \\
& 256 & 0.1018 & 0.1392 & 0.0628 &-0.0017 & 0.2960 & 0.1543 & 0.0762 & 0.0758 & 0.4203 & 0.4399 & 0.1307 & 0.1077 \\
& 512 & 0.1097 & 0.1385 & -0.0048 &-0.0009 & 0.3392 & 0.1337 & 0.0018 & 0.0214 & 0.4852 & 0.4943 & 0.1338 & 0.2019 \\
& 768 & 0.0937 & 0.1192 & 0.0145 & 0.0416 & 0.2732 & 0.0460 & -0.0179 & 0.0195 & 0.5522 & 0.5970 & 0.0702 & 0.1630 \\
& 1024 & 0.1283 & 0.1432 & -0.0370 &-0.0340 & 0.3719 & 0.2157 & -0.0342 & 0.0083 & 0.6066 & 0.5695 & 0.1325 & 0.1403 \\
& 1536 & 0.0085 & -0.0191 & -0.0914 &-0.1322 & 0.3975 & 0.2347 & -0.0684 & -0.0904 & 0.6035 & 0.6215 & 0.1883 & 0.4055 \\
& 2048 &-0.0135 & 0.0233 & -0.0181 &-0.0131 & 0.3929 & 0.1843 & -0.1325 & -0.1087 & 0.5058 & 0.4803 & 0.2679 & 0.3719 \\
& 4096 & 0.1096 & 0.2106 & -0.0675 &-0.1011 & 0.3472 & 0.2168 & -0.0838 & -0.1240 & 0.4476 & 0.4480 & 0.3188 & 0.3158 \\\midrule
\multirow{8}{*}{NLI} & 128 &-0.0260 & -0.0689 & 0.0117 & 0.0824 & 0.3635 & 0.2411 & 0.0086 & -0.0107 & 0.5041 & 0.5647 & 0.1202 & 0.2608 \\
& 256 & 0.0152 & -0.0043 & -0.0119 & 0.0548 & 0.2937 & 0.1005 & -0.0263 & -0.0365 & 0.4199 & 0.3586 & 0.0890 & 0.1729 \\
& 512 & 0.0841 & 0.0836 & 0.0434 & 0.0034 & 0.3480 & 0.2177 & -0.0558 & -0.0369 & 0.4783 & 0.4905 & 0.1185 & 0.1280 \\
& 768 & 0.0651 & 0.0741 & -0.0624 &-0.0847 & 0.3491 & 0.0833 & 0.0128 & 0.0177 & 0.3564 & 0.4090 & 0.2651 & 0.3405 \\
& 1024 & 0.0769 & 0.0800 & -0.0105 &-0.0207 & 0.3813 & 0.1694 & 0.0212 & 0.0397 & 0.5264 & 0.5492 & 0.0781 & 0.1539 \\
& 1536 & 0.0986 & 0.0605 & -0.0190 &-0.0318 & 0.4322 & 0.3107 & -0.1126 & -0.0961 & 0.5368 & 0.5467 & 0.0161 & 0.2438 \\
& 2048 & 0.0839 & 0.0725 & -0.0183 & 0.0097 & 0.4139 & 0.2372 & -0.0292 & -0.0113 & 0.5071 & 0.5701 & -0.1031 & 0.1544 \\
& 4096 & 0.0493 & 0.0783 & -0.0033 & 0.0081 & 0.4562 & 0.3065 & -0.0401 & -0.0502 & 0.4496 & 0.4980 & 0.1686 & 0.1988 \\\midrule
Full & - & 0.0786 & 0.1205 & 0.2994 & 0.3551 & -0.0173 & -0.0144 & 0.0344 & -0.0107 & 0.4904 & 0.4617 & 0.1397 & 0.1489\\\midrule
\rowcolor{gray!9}
Full (GPT-4) & - & 0.6078 & 0.4561 & 0.325 & 0.3404 & 0.5801 & 0.5185 & 0.1197 & 0.1061 & 0.6352 & 0.6964 & 0.5119 & 0.3758\\
\bottomrule
\end{tabular}
}
\caption{All results of correlation with human evaluations by \texttt{gpt-3.5-turbo-16k-0613}.}
\label{tab:full_table_gpt3.5}
\end{table*}

\clearpage

\section{Analysis of source document length distribution under various length limitations}
We evaluated the length distribution of the extracted source documents across various length limitations. As illustrated in Table~\ref{tab:length_dist}, there is generally no significant difference in length distribution under different length limitations, suggesting minimal information loss. However, an exception is observed when the length limitation is set to a longer value, such as 4096 tokens. This discrepancy is attributable to some original source documents being shorter than 4096 tokens, which influences the average length due to the presence of these shorter instances.

\begin{table*}[ht]
    \centering
    \resizebox{\textwidth}{!}{
\begin{tabular}{lcrrrrrrrrrrrr}
\toprule
    & & \multicolumn{3}{c}{\textbf{arXiv}} & \multicolumn{3}{c}{\textbf{GovReport}} & \multicolumn{3}{c}{\textbf{PubMed}} & \multicolumn{3}{c}{\textbf{SQuALITY}} \\
    \cmidrule(lr){3-5} \cmidrule(lr){6-8} \cmidrule(lr){9-11} \cmidrule(lr){12-14}
    \textbf{Methods}    & \textbf{Length}     &  \textbf{avg.} &  \textbf{25\%} &  \textbf{75\%} &  \textbf{avg.} &  \textbf{25\%} &  \textbf{75\%} &  \textbf{avg.} &  \textbf{25\%} &  \textbf{75\%} &  \textbf{avg.} &  \textbf{25\%} &  \textbf{75\%} \\
\midrule
\multirow{8}{*}{LEAD} & 128  &       108.8 &      105.0 &      116.0 &      98.5 &     93.0 &    112.0 &         94.6 &        84.8 &       116.2 &          112.3 &         108.8 &         119.2 \\
        & 256  &       223.5 &      217.0 &      228.0 &     227.6 &    218.0 &    239.0 &        228.3 &       220.5 &       237.0 &          233.0 &         229.0 &         237.2 \\
        & 512  &       477.6 &      472.0 &      488.0 &     474.1 &    461.0 &    490.0 &        475.0 &       466.2 &       486.8 &          475.6 &         471.0 &         480.2 \\
        & 768  &       722.5 &      719.0 &      732.0 &     727.9 &    718.0 &    738.0 &        709.0 &       675.5 &       733.2 &          712.3 &         701.2 &         725.5 \\
        & 1024 &       970.7 &      961.0 &      982.0 &     969.4 &    958.0 &    987.0 &        974.9 &       967.0 &       983.2 &          954.6 &         950.8 &         962.0 \\
        & 1536 &     1,456.5 &    1,448.0 &    1,467.0 &   1,457.9 &  1,449.0 &  1,469.0 &      1,450.0 &     1,450.0 &     1,480.2 &        1,433.9 &       1,411.8 &       1,448.2 \\
        & 2048 &     1,921.1 &    1,939.0 &    1,960.0 &   1,963.4 &  1,955.0 &  1,976.0 &      1,889.5 &     1,927.5 &     1,973.0 &        1,916.1 &       1,894.0 &       1,939.5 \\
        & 4096 &     3,639.1 &    3,886.0 &    3,943.0 &   3,752.1 &  3,634.0 &  3,965.0 &      3,015.2 &     2,297.8 &     3,917.2 &        3,834.0 &       3,795.0 &       3,882.2 \\\midrule
\multirow{8}{*}{ROUGE-1} & 128  &       103.7 &       95.8 &      122.0 &      64.5 &      0.0 &    103.0 &         85.6 &        70.2 &       111.5 &           96.2 &          83.0 &         115.2 \\
        & 256  &       239.5 &      232.8 &      250.0 &     226.4 &    208.0 &    243.0 &        226.6 &       220.2 &       244.2 &          236.5 &         227.8 &         248.0 \\
        & 512  &       491.6 &      486.0 &      501.0 &     478.0 &    466.0 &    499.0 &        488.1 &       477.0 &       501.0 &          497.0 &         489.0 &         506.2 \\
        & 768  &       746.8 &      741.0 &      758.0 &     739.5 &    732.0 &    754.0 &        740.6 &       729.0 &       756.0 &          757.5 &         752.8 &         764.0 \\
        & 1024 &     1,005.6 &      999.0 &    1,015.0 &     999.8 &    990.8 &  1,014.0 &      1,001.4 &       994.0 &     1,016.2 &        1,015.4 &       1,010.5 &       1,020.2 \\
        & 1536 &     1,511.2 &    1,505.0 &    1,524.0 &   1,511.2 &  1,504.0 &  1,524.0 &      1,486.8 &     1,491.8 &     1,519.0 &        1,529.6 &       1,524.8 &       1,538.2 \\
        & 2048 &     1,990.8 &    2,010.8 &    2,035.0 &   2,021.1 &  2,012.8 &  2,035.0 &      1,942.2 &     2,000.8 &     2,030.0 &        2,047.3 &       2,041.8 &       2,055.0 \\
        & 4096 &     3,739.2 &    4,025.5 &    4,072.0 &   3,822.1 &  3,634.0 &  4,073.2 &      3,046.9 &     2,297.8 &     4,014.2 &        4,109.4 &       4,093.0 &       4,121.0 \\\midrule
\multirow{8}{*}{ROUGE-2} & 128  &       113.0 &      106.0 &      122.0 &      82.8 &     71.8 &    114.0 &         96.5 &        91.8 &       116.5 &          107.8 &         103.8 &         123.0 \\
        & 256  &       236.4 &      228.0 &      247.0 &     224.2 &    212.8 &    243.0 &        224.1 &       215.2 &       242.0 &          241.3 &         231.0 &         250.2 \\
        & 512  &       492.5 &      487.0 &      504.0 &     482.7 &    472.0 &    500.2 &        480.1 &       471.0 &       494.5 &          496.6 &         487.0 &         506.0 \\
        & 768  &       747.9 &      741.0 &      758.0 &     740.7 &    733.0 &    756.2 &        738.8 &       731.2 &       756.0 &          755.1 &         751.0 &         762.2 \\
        & 1024 &     1,002.7 &      994.0 &    1,014.0 &     994.6 &    983.5 &  1,012.0 &      1,000.6 &       996.0 &     1,017.0 &        1,012.9 &       1,007.5 &       1,021.2 \\
        & 1536 &     1,509.7 &    1,503.0 &    1,522.0 &   1,511.6 &  1,504.0 &  1,524.0 &      1,492.1 &     1,500.8 &     1,527.0 &        1,530.0 &       1,522.8 &       1,538.0 \\
        & 2048 &     1,991.0 &    2,015.0 &    2,033.0 &   2,015.5 &  2,015.0 &  2,033.2 &      1,945.8 &     2,002.0 &     2,031.0 &        2,049.2 &       2,043.8 &       2,056.0 \\
        & 4096 &     3,739.2 &    4,025.5 &    4,072.0 &   3,822.1 &  3,634.0 &  4,073.2 &      3,046.9 &     2,297.8 &     4,014.2 &        4,109.4 &       4,093.0 &       4,121.0 \\\midrule
\multirow{8}{*}{ROUGE-1+2} & 128  &       108.2 &      101.8 &      122.0 &      75.7 &     61.5 &    109.0 &         95.0 &        90.5 &       119.0 &          100.0 &          93.8 &         117.2 \\
        & 256  &       238.5 &      232.0 &      249.0 &     225.0 &    206.0 &    244.2 &        225.4 &       215.0 &       242.5 &          240.6 &         234.5 &         250.0 \\
        & 512  &       491.3 &      484.0 &      501.2 &     479.0 &    467.0 &    499.0 &        485.3 &       477.0 &       502.2 &          498.6 &         492.8 &         505.2 \\
        & 768  &       747.3 &      740.8 &      760.0 &     741.6 &    728.8 &    757.0 &        736.1 &       726.8 &       751.5 &          755.2 &         746.8 &         763.2 \\
        & 1024 &     1,004.2 &      996.0 &    1,014.0 &     996.6 &    988.0 &  1,012.2 &        997.0 &       988.5 &     1,015.2 &        1,016.2 &       1,012.5 &       1,021.2 \\
        & 1536 &     1,511.1 &    1,502.8 &    1,524.0 &   1,506.4 &  1,498.0 &  1,522.0 &      1,482.8 &     1,491.2 &     1,522.2 &        1,530.3 &       1,524.0 &       1,536.8 \\
        & 2048 &     1,989.5 &    2,011.0 &    2,032.2 &   2,022.6 &  2,014.0 &  2,035.2 &      1,938.7 &     1,990.2 &     2,026.0 &        2,047.1 &       2,041.5 &       2,052.2 \\
        & 4096 &     3,739.2 &    4,025.5 &    4,072.0 &   3,822.1 &  3,634.0 &  4,073.2 &      3,046.9 &     2,297.8 &     4,014.2 &        4,109.4 &       4,093.0 &       4,121.0 \\\midrule
\multirow{8}{*}{BERTScore} & 128  &       109.7 &      101.0 &      122.0 &      77.5 &     67.2 &    112.2 &         90.0 &        87.0 &       111.0 &          110.2 &         113.2 &         125.0 \\
        & 256  &       237.6 &      226.0 &      248.2 &     232.9 &    219.0 &    246.0 &        221.3 &       203.2 &       240.0 &          243.0 &         236.8 &         252.2 \\
        & 512  &       483.7 &      475.0 &      502.0 &     490.5 &    481.0 &    504.0 &        472.9 &       453.0 &       498.5 &          503.0 &         497.8 &         510.0 \\
        & 768  &       749.8 &      738.0 &      758.0 &     746.7 &    742.0 &    756.0 &        736.4 &       718.8 &       753.0 &          759.6 &         751.8 &         769.0 \\
        & 1024 &       997.3 &      989.8 &    1,012.0 &   1,001.0 &    993.8 &  1,013.0 &        990.2 &       976.8 &     1,007.5 &        1,019.1 &       1,014.0 &       1,021.0 \\
        & 1536 &     1,511.4 &    1,501.0 &    1,524.2 &   1,513.2 &  1,503.8 &  1,526.0 &      1,488.7 &     1,497.8 &     1,518.5 &        1,532.5 &       1,525.8 &       1,543.2 \\
        & 2048 &     1,988.9 &    2,014.0 &    2,034.2 &   2,023.0 &  2,013.0 &  2,036.0 &      1,945.5 &     1,999.8 &     2,031.2 &        2,047.0 &       2,040.0 &       2,055.2 \\
        & 4096 &     3,736.2 &    3,947.2 &    4,074.0 &   3,823.7 &  3,634.0 &  4,076.0 &      3,048.0 &     2,297.8 &     4,035.8 &        4,107.4 &       4,092.5 &       4,119.0 \\\midrule
\multirow{8}{*}{NLI} & 128  &       105.9 &       97.0 &      116.0 &     107.0 &    100.8 &    115.2 &        100.4 &        93.0 &       117.5 &          110.7 &         105.8 &         116.0 \\
        & 256  &       229.6 &      222.0 &      240.0 &     230.3 &    223.0 &    239.2 &        228.9 &       224.8 &       238.5 &          228.4 &         225.2 &         233.2 \\
        & 512  &       472.7 &      466.0 &      484.0 &     473.3 &    465.0 &    483.0 &        471.8 &       460.8 &       485.2 &          466.3 &         460.0 &         474.0 \\
        & 768  &       719.9 &      711.0 &      731.0 &     720.3 &    711.0 &    731.0 &        720.7 &       717.5 &       737.5 &          707.5 &         700.5 &         715.2 \\
        & 1024 &       962.3 &      957.8 &      977.0 &     966.7 &    956.8 &    980.0 &        973.8 &       968.8 &       988.2 &          946.1 &         938.0 &         958.0 \\
        & 1536 &     1,456.1 &    1,446.0 &    1,471.0 &   1,460.7 &  1,450.0 &  1,475.0 &      1,444.8 &     1,454.0 &     1,476.2 &        1,426.4 &       1,415.5 &       1,442.2 \\
        & 2048 &     1,924.1 &    1,930.8 &    1,960.0 &   1,954.0 &  1,943.0 &  1,970.0 &      1,895.0 &     1,936.0 &     1,974.0 &        1,905.6 &       1,896.0 &       1,922.0 \\
        & 4096 &     3,637.2 &    3,875.0 &    3,942.2 &   3,736.6 &  3,634.0 &  3,953.2 &      3,013.2 &     2,297.0 &     3,915.5 &        3,827.2 &       3,801.5 &       3,865.0 \\
\bottomrule
\end{tabular}
}
\caption{Distribution of source document lengths under different length limitations.}
\label{tab:length_dist}
\end{table*}

\clearpage

\section{Dataset license}
Table~\ref{tab:license} provides a summary of the licenses associated with datasets used in this work.

\begin{table}[h!]
    \centering
    \resizebox{\textwidth}{!}{
    \begin{tabular}{cccc}
    \toprule
        Data & Data License & Annotation & Annotation License\\\midrule
        arXiv~\cite{cohan2018discourse} & Apache License 2.0 & \citet{koh-etal-2022-far} & Unspecified\\
        GovReport~\cite{huang2021efficient} & Unspecified & \citet{koh-etal-2022-far} & Unspecified\\
        PubMed~\cite{cohan2018discourse} & Apache License 2.0 & \citet{krishna-etal-2023-longeval} & Apache License 2.0\\
        SQuALITY~\cite{wang2022squality} & Unspecified & \citet{krishna-etal-2023-longeval} & Apache License 2.0\\
    \bottomrule
    \end{tabular}
    }
    \caption{Summary of dataset licenses.}
    \label{tab:license}
\end{table}

\section{The design choice of LLM-based evaluator}
\label{sec:design_choice}
In our preliminary experiments, we attempted to conduct summary evaluation using the prompting approach based on the G-Eval setting~\cite{liu2023geval}, which sets the \texttt{temperature} parameter to \texttt{1} and the number of completions \texttt{n} to \texttt{20}. However, when we applied this approach to the long-document summarization evaluation dataset, we encountered a "Rate limit issue." Since we did not encounter this error when we set the parameter \texttt{n} to \texttt{1}, we suspect it may be an issue on the API side.

As an alternative method, we considered making 20 API calls to obtain 20 samples. However, this could lead to a 20-fold increase in the cost of evaluating a single instance, which is not a practical solution, even though the original pricing formula is \texttt{num\_tokens(input) + max\_tokens * max(n, best\_of)}.\footnote{\url{https://openai.com/pricing}}

In addition to this, we conducted further preliminary experiments in the benchmark for short-text summarization evaluation using the SummEval dataset~\cite{fabbri2021summeval}. Specifically, we performed sub-sampling to create a smaller subset of the dataset and conducted summary evaluations in two settings: the original G-Eval setting with \texttt{temperature = 1} and \texttt{n = 20}, and a deterministic setting\footnote{Theoretically speaking, a language model with a \texttt{temperature} setting of \texttt{0} should produce deterministic output. However, it is known that GPT-4 can still generate random outputs even when the \texttt{temperature} is set to \texttt{0}. Nevertheless, in our specific setup, where the output is limited to a single token and unlike typical text generation problems, error propagation is not a concern. In fact, when we set the temperature to 0 and generated output 10 times for 10 different instances, we observed that in one instance, 7 out of 10 times, it was estimated to be 5, and 3 out of 10 times, it was estimated to be 4. In other words, we found that deterministic inference was possible approximately 97\% of the time.} with \texttt{temperature = 0} and \texttt{n = 1}. This small study revealed that we obtained nearly identical results in both cases.

Based on these observations, in our main experiments, we evaluated the summaries with \texttt{temperature = 0}, which allowed us to achieve \textit{relatively} higher reproducibility of results compared to the original setting without facing "Rate limit issue".

\section{Additional results}
\label{sec:additional_results}
\begin{figure*}[!ht]
    \centering
    \includegraphics[width=\linewidth]{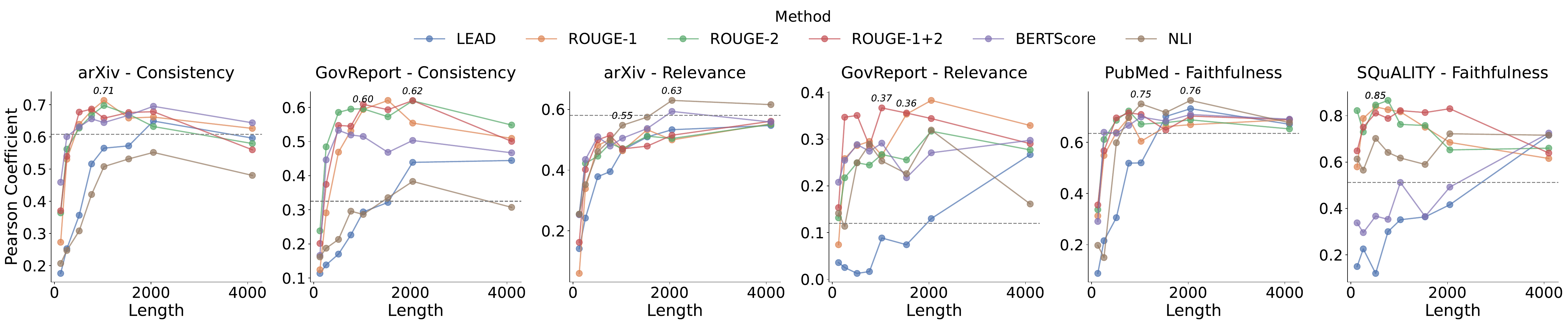}
    \caption{Relationship between document length and Pearson correlation}
    \label{fig:pearson_appdx}
    \postspace
\end{figure*}

\begin{figure*}[!ht]
    \centering
    \includegraphics[width=\linewidth]{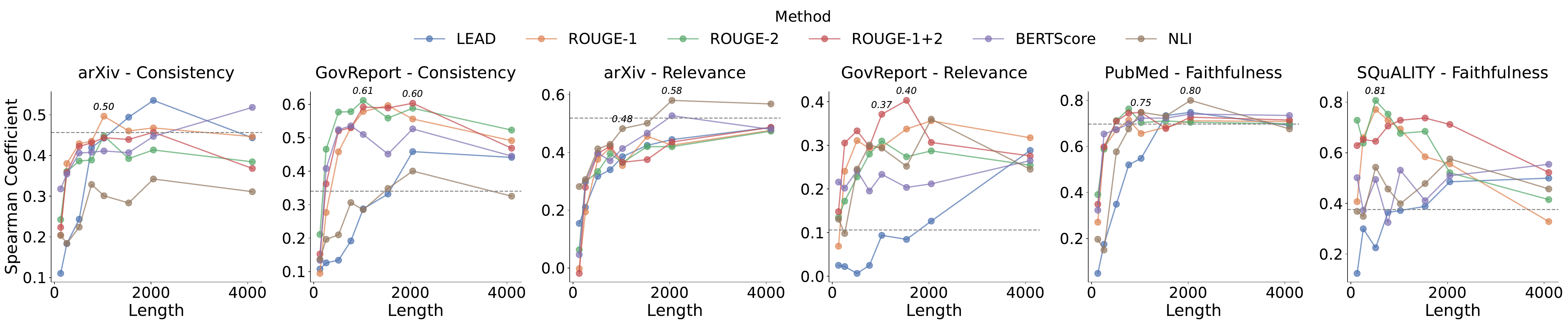}
    \caption{Relationship between document length and Spearman's rank correlation.}
    \label{fig:spearman}
\end{figure*}

We show the same plot as shown in Figure~\ref{fig:pearson} (Figure ~\ref{fig:pearson_appdx} repeats here for convenience of readers), but we use Spearman's rank correlation instead of Pearson's in Figure~\ref{fig:spearman}. The observation is nearly the same as in the Pearson case.

\section{SQuALITY dataset issue}
\label{sec:squality_issue}
We conducted experiments using manually annotated human scores for the SQuALITY dataset by \citet{krishna-etal-2023-longeval}. However, in our preliminary experiments, we observed significant differences in correlation when using baseline metrics, such as ROUGE-1 F1 scores, compared to those reported in the paper.

\begin{figure}[!ht]
    \centering
    \includegraphics[width=0.7\linewidth]{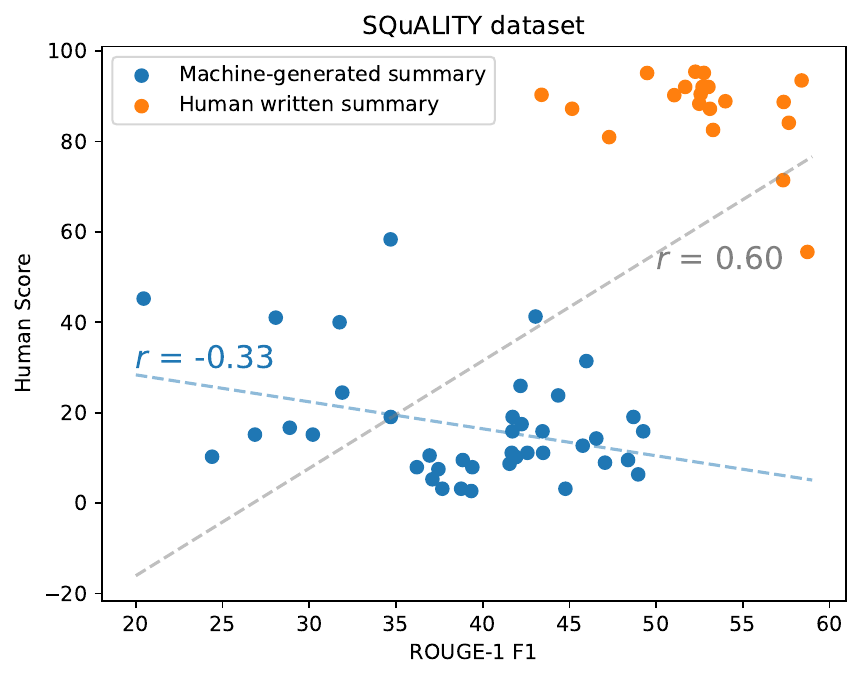}
    \caption{The relationship between the ROUGE-1 F1 score and the human score with or without including human-written summaries for correlation calculation}
    \label{fig:squality_issue}
\end{figure}

Upon closer examination, we discovered that \citet{krishna-etal-2023-longeval} used reference summaries to compute correlations in the SQuALITY dataset. As depicted in Figure~\ref{fig:squality_issue}, the reference summary ({\color{orange}orange dot}) is generally evaluated as faithful, resulting in excessively high automatic evaluation scores and a correlation of $r = 0.6$.

In fact, when we re-evaluated the correlation between the ROUGE-1 F1 score and the human scores without using human-written summaries ({\color{bblue}blue dot}), we found a significant drop in correlation to $r = -0.33$. Therefore, the results presented in Table~\ref{tab:main_result} are inconsistent with those reported in the original paper~\cite{krishna-etal-2023-longeval}.

\section{Relevant Work}
\paragraph{Evaluation of Text Generation:}
Evaluation of text generation plays a critical role in the development of high-quality text generation systems~\cite{celikyilmaz2021evaluation}. However, most automatic evaluation metrics do not always correlate well with human evaluation~\cite{kryscinski2019evaluating,bhandari-etal-2020-evaluating,fabbri2021summeval,adams2023meta}. 
Recently, LLMs have shown a strong alignment with human judgment for the evaluation of text generation~\cite{chiang-lee-2023-large,liu2023geval,fu2023gptscore}. Still, LLMs are computationally expensive, meaning that long document summary evaluation can be costly.
Our study shows that extracting important sentences in advance not only reduces inference costs but also exhibits a higher correlation with human evaluations.

\paragraph{NLP for Long Sequence:}
NLP studies have begun to shift from focusing on individual sentences to long documents.
In particular, there has been a lot of effort in developing Transformer models that can effectively analyze longer sequences~\cite{beltagy2020longformer,gu2022efficiently,dao2022flash}. 
However, such models often perform poorly when important information is in the middle~\cite{liu2023lost}.
Our study identified a similar problem with long document summary evaluation and introduced a cost-effective solution.

\end{document}